\documentclass{article}

\usepackage{amsmath,amsfonts,bm}

\def\eqref#1{equation~\ref{#1}}

\def\1{\bm{1}}
\newcommand{\train}{\mathcal{D_{\mathrm{train}}}}
\newcommand{\valid}{\mathcal{D_{\mathrm{valid}}}}

\DeclareMathAlphabet{\mathsfit}{\encodingdefault}{\sfdefault}{m}{sl}
\SetMathAlphabet{\mathsfit}{bold}{\encodingdefault}{\sfdefault}{bx}{n}

\usepackage{microtype}
\usepackage{graphicx}
\usepackage{booktabs} 
\usepackage{amsmath}
\usepackage{caption}
\usepackage{listings}
\usepackage{multirow}
\usepackage{subcaption}
\usepackage{longtable}

\usepackage{hyperref}

\usepackage[accepted]{icml2024}

\usepackage{amsmath}
\usepackage{amssymb}
\usepackage{mathtools}
\usepackage{amsthm}
\usepackage{listings}

\theoremstyle{plain}

\theoremstyle{definition}

\theoremstyle{remark}

\usepackage[textsize=tiny]{todonotes}

\icmltitlerunning{CodeIt: Self-Improving Language Models with Prioritized Hindsight Replay} 

\begin{document}

\twocolumn[
\icmltitle{CodeIt: Self-Improving Language Models with Prioritized Hindsight Replay} 



\icmlsetsymbol{equal}{*}

\begin{icmlauthorlist}
    \icmlauthor{Natasha Butt}{uva,qcom}
    \icmlauthor{Blazej Manczak}{qcom}
    \icmlauthor{Auke Wiggers}{qcom} \\
    \icmlauthor{Corrado Rainone}{qcom}
    \icmlauthor{David W. Zhang}{qcom}
    \icmlauthor{Michaël Defferrard}{qcom}
    \icmlauthor{Taco Cohen}{qcom,emp}
\end{icmlauthorlist}

\icmlaffiliation{uva}{University of Amsterdam}
\icmlaffiliation{qcom}{Qualcomm AI Research. Qualcomm AI Research is an initiative of Qualcomm Technologies, Inc.}
\icmlaffiliation{emp}{Work was completed while an employee at Qualcomm Technologies Netherlands B.V.}

\icmlcorrespondingauthor{Natasha Butt}{n.e.butt@uva.nl}

\icmlkeywords{Machine Learning, ICML}

\vskip 0.3in
]

\printAffiliationsAndNotice{} 

\begin{abstract}
    Large language models are increasingly solving tasks that are commonly believed to require human-level reasoning ability.
However, these models still perform very poorly on benchmarks of general intelligence such as the Abstraction and Reasoning Corpus (ARC).
In this paper, we approach ARC as a programming-by-examples problem, and introduce a novel and scalable method for language model self-improvement called Code Iteration (CodeIt).
Our method iterates between
1) program sampling and hindsight relabeling, and
2) learning from prioritized experience replay.
By relabeling the goal of an episode (i.e., the target program output given input) to the realized output produced by the sampled program, our method effectively deals with the extreme sparsity of rewards in program synthesis.
Applying CodeIt to the ARC dataset, we demonstrate that prioritized hindsight replay, along with pre-training and data-augmentation, leads to successful inter-task generalization.
CodeIt is the first neuro-symbolic approach that scales to the full ARC evaluation dataset.
Our method solves 15\% of ARC evaluation tasks, achieving state-of-the-art performance and outperforming existing neural and symbolic baselines.
Our code is available at \url{https://github.com/Qualcomm-AI-research/codeit}.

\end{abstract}

\section{Introduction}



The Abstraction and Reasoning Corpus (ARC) is a general artificial intelligence benchmark targeted at both humans and AI systems \cite{chollet2019measure}.
ARC is a challenging benchmark because it contains few-shot example tasks that assume access to the
four innate core knowledge systems: objects, actions, number, and space \cite{knowledge}.
It was designed to require no knowledge outside of these priors, and so the massive memorization capability of pre-trained language models is of limited use for this problem.
Humans are able to solve 80\% of (a random subset of) ARC tasks in user studies \cite{human_arc}, 
whereas state-of-the-art neural approaches based on GPT-4 solve only 12\% of evaluation tasks \cite{gendron2023large}.

\begin{figure}[t]
    \centering
    \includegraphics[width=0.49\textwidth,trim={0 4mm 0 0}]{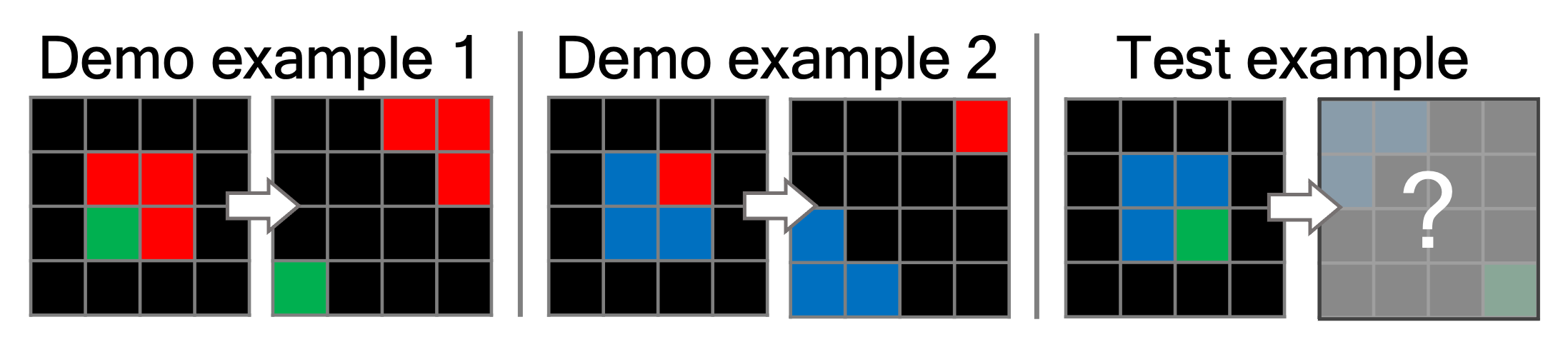}
    \caption{A simplified ARC task. Given two demonstration input-output pairs, the goal is to determine the output grid for the test example, in three attempts or fewer. 
    The size of the grids and the number of demonstration and test examples differs across tasks.}
    \label{fig:method:arcexample}
    \vspace{-1mm}
\end{figure}

Each ARC task consists of a number of \emph{demonstration examples}, each consisting of an input and output grid, and one or more test inputs for which the corresponding output must be predicted (see ~\autoref{fig:method:arcexample}).
Effective agents use abstractions related to the four core knowledge systems, generalize from demonstration to test examples, and generalize between tasks. For example, an agent may infer that adjacent cells (space) of the same color value (number) form an object. An agent may also infer that multiple objects sometimes attract or repel (action). Using these abstractions to reason about the value of the test output, an agent may generalize from the demonstration examples to the test example.

\begin{figure}
    \centering
    \includegraphics[width=0.49\textwidth, trim={0 7mm 0 0}]{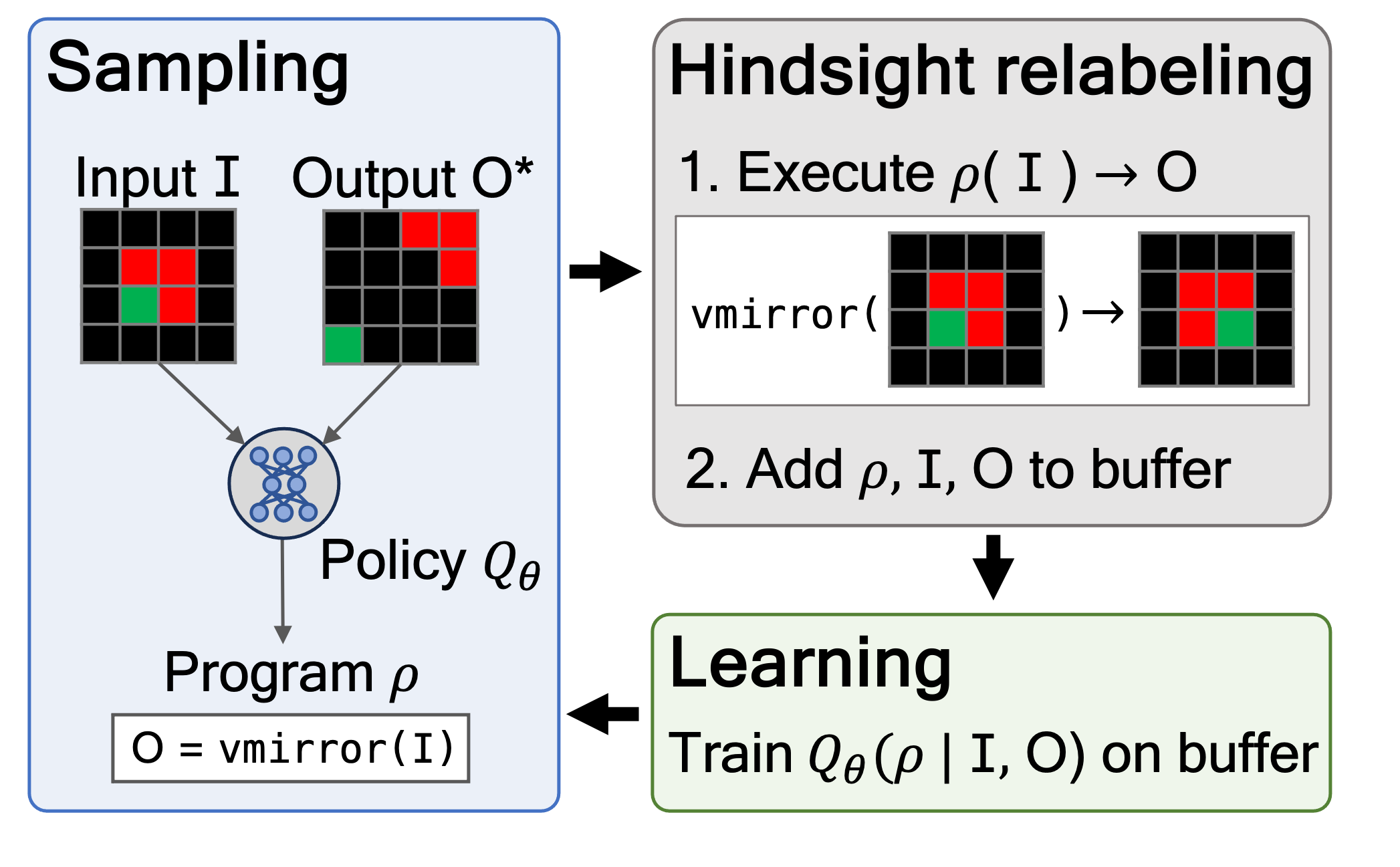}
    \caption{An overview of Code Iteration. In the sampling stage, programs $\rho$ are sampled from the policy $Q_\theta$ conditioned on input-output pairs. 
    The program may not produce target output $O^*$ given $I$, so we use hindsight relabeling: we execute the program, and add the program $\rho$, inputs $I$, and \emph{realized} outputs $O$ to the buffer. 
    In the learning stage, we train the policy on samples from the buffer.}
    \label{fig:overview}
    \vspace{-1mm}
\end{figure}

Existing approaches to ARC can be classified as either neural \cite{gendron2023large, mirchandani2023large}, meaning they directly predict output grids using a neural network, or (neuro-) symbolic \citep{arc_vision_image,ferre2021first, ferre2023tackling}, meaning they first predict a program or other symbolic representation of the mapping between input and output grids, before using it to generate the output grids.
Through the use of a well-designed \emph{domain-specific language} (DSL), the symbolic methods can be endowed with prior knowledge analogous to the core knowledge systems found in humans.
By combining neural networks and symbolic representations like programs, the system can leverage both prior knowledge and data to solve the ARC tasks.



However, the most effective existing methods, whether neural or symbolic, fail to use experience to generalize between tasks. 
We propose using Expert Iteration (ExIt) \citep{Anthony2017-hb} to incorporate experience. 
ExIt methods do this by alternating between two phases: gathering data with an (often expensive) exploration policy, and improving the policy by training on the newfound experiences. 
Instead of performing ExIt in the grid space, we take a neuro-symbolic approach and train our model to learn to write programs.
This brings us closer to the system that emulates general fluid intelligence described by \citet{chollet2019measure}: by incorporating new experiences in the form of abstractions. 
Recent ExIt approaches employ self-improving language models \cite{gulcehre2023reinforced, aksitov2023rest, wang2023enable} to replace the expensive expert by sampling from a language model policy and reward-based filtering, saving only 
trajectories that obtain high reward. 
This allows them to scale well and benefit from knowledge already captured in the policy.
These methods prove effective on program synthesis tasks with natural language specifications \cite{singh2023human} and code specifications \cite{haluptzok2022language}. 
However, when solving ARC, agents start ExIt with poor prior knowledge about the search space, as the task is out-of-distribution. 
Finding a correct program is challenging: positive rewards are extremely sparse.
As a result, these methods are sample inefficient in the context of ARC, and programming-by-examples more generally. 
To enable learning in sparse-reward settings, hindsight relabeling \cite{andrychowicz2017hindsight} creates artificial expert trajectories post-hoc,
and methods that combine ExIt and this technique
have improved sample efficiency \cite{gauthier2022program, butt2022program}. 
However, since the relabelled data distribution is constantly changing, there is risk of catastrophic forgetting \cite{catastrophic}.



In this work, we introduce a novel, scalable expert iteration method for sparse reward settings that does not suffer from catastrophic forgetting.
Our method, which we call Code Iteration or \emph{CodeIt} for short,  iterates between 1) a sampling and hindsight relabeling stage and 2) a learning stage with prioritized experience replay. 
We show a visualization in \autoref{fig:overview}.
This iterative procedure thus allows us to automatically generate new data without human intervention. 
Unlike current self-improvement approaches that perform sampling and filtering \cite{singh2023human}, CodeIt learns from all program samples, improving sample efficiency. 
By prioritizing training on experiences that solve real tasks, we ameliorate the risk of catastrophic forgetting.


CodeIt solves 59/400 ARC evaluation tasks, achieving state-of-the-art performance by learning from experiences in the form of abstractions and generalizing to new tasks.
We analyze the programs discovered by CodeIt and find that these are on average shorter and use different primitives compared to our custom symbolic baselines. 
Furthermore, after finding an initial solution, CodeIt continues to improve it over time; shorter solutions are found in 53\% of solved ARC tasks,
highlighting the ability to perform program refinement.
We perform careful ablations to better understand the impact on task performance of key components: ExIt, prioritized hindsight replay, and prior knowledge.  

\section{Method}

We approach ARC as a programming-by-examples problem: for a given set of tasks that we call the \emph{search set}, we aim to find programs that correctly match inputs with their respective outputs, and we do so by training a \emph{policy} to produce programs when shown demonstration examples. 
This is achieved by iterating between two stages: 1) writing programs using a policy and applying hindsight relabeling, and 2) learning from the programs and their input-output examples. 
We first describe key design choices below, and then explain the iterative procedure.

\subsection{Design choices}

\paragraph{Programming language}  

We restrict our programming language to the open source domain specific language (DSL) of \citet{hodel}.
Although various open source DSLs for ARC exist, Hodel designed their DSL using only the ARC training split, whereas some authors incorporate priors from the ARC evaluation split into their DSLs \cite{kaggle}.

Hodel's DSL contains grid manipulation functions (e.g., \texttt{vmirror} or \texttt{hmirror}, which mirror the grid along the vertical or horizontal axis), \texttt{fill} functions that replace all pixels of a certain color, and functions that return locations of specific pixel groups. 
See Appendix \ref{appendix:hodeldsl} for details on the DSL and more example primitives, and see \citet{hodel} for discussion on the DSL's primitives and capability.

\paragraph{Policy}

Our choice of policy is a pretrained encoder-decoder Large Language Model (LLM).
We use the 220 million parameter CodeT5+ \citep{wang2023codet5plus} model and its default tokenizer, which are pretrained on a diverse set of programming tasks.
We input the demonstration examples to the encoder, and let the decoder generate the corresponding program.
If necessary, demonstration examples are truncated to fit in the encoder context window.

\begin{figure}
    \centering
    \includegraphics[width=0.46\textwidth]{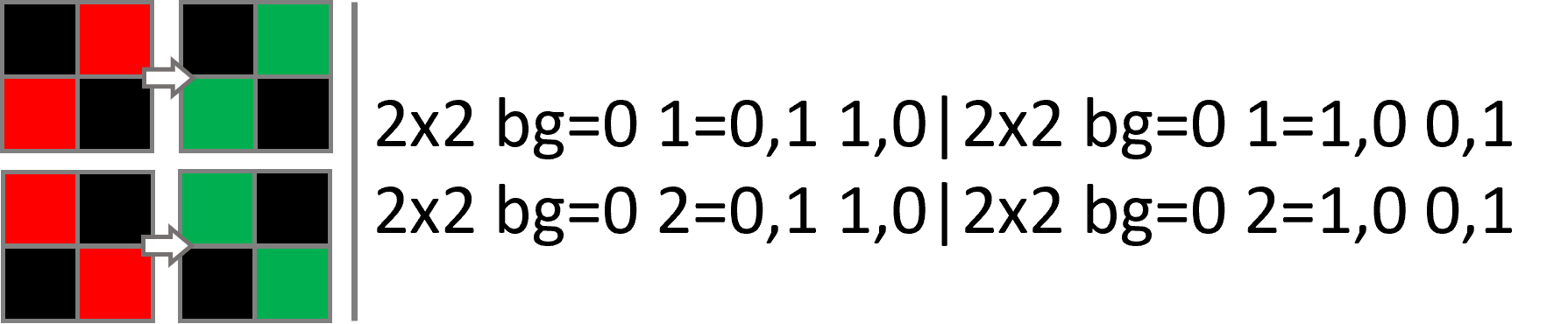}
    \caption{Sparse grid representation of a simplified ARC task.}
    \label{fig:sparse_representation}
\end{figure}

\paragraph{Grid representation}
In order to condition the language model policy on input-output grids, we represent them as text.
Instead of encoding the grid as a 2-dimensional array, we use an object-centric text representation.
Each color is encoded as an integer, and for each color in the grid we list all the grid cells with that color as $[x,y]$ coordinates.  
Since the majority of cells belong to the background color, this procedure significantly reduces the number of tokens required to encode the grid (see Figure~\ref{fig: grid_tokens} in Appendix \ref{subsec:task_representation}).
An example of the sparse grid representation is shown in Figure \ref{fig:sparse_representation}.
This object-centric text representation, similar to the one of \citet{xu2023llms}, works well for sparse grids and is human-interpretable.

\subsection{The Code Iteration Algorithm}

We initialize the policy network by training on ground truth data.
We then start CodeIt, iterating between \textit{sampling and hindsight relabeling} and \textit{learning}.
We refer to one full pass of sampling and learning as a \emph{meta-iteration}.
We show the procedure in Fig.~\ref{fig:overview}, and explain each stage in more detail below.
For pseudocode, see Appendix \ref{subsec: pseudo}.

\paragraph{Initialization}

We start from a dataset of ARC training tasks and solution programs written in the domain-specific language (DSL) of \citet{hodel}, which we call the \emph{training set}.
This dataset is expanded by randomly mutating programs (for details of this procedure, see 
Appendix \ref{subsec:mutation}), resulting in an \emph{augmented training set}.

The initial dataset augmentation step serves multiple purposes.
Mixing in mutated programs acts as a form of data augmentation, and is a common approach in policy improvement for program synthesis \citep{ellis2020dreamcoder, alphatensor}.
Before experiences are sampled from the policy, the model can already learn the DSL syntax, which can be challenging if the training set is small.
It also enables the model to learn how to interpret the task demonstration examples before we begin iterative learning, improving the quality of our policy samples in early meta-iterations.

\paragraph{Sampling and hindsight relabeling}\label{ssec:sampling stage}

In the sampling stage, we obtain new programs using the policy $Q_\theta$.
Let the \textit{search set} be the set of tasks for which we want to find a corresponding program. 
For each task in the search set, we convert the demonstration examples' input $I$ and target output $O^*$ from grid to text representation, encode these using the policy, and then autoregressively decode a program:
$ \rho \sim Q_\theta( \rho | I , O^*)$.
We then run the obtained program on the input grids.
If the program is syntactically incorrect or the runtime is too high, we discard it.
Otherwise, we obtain program outputs $O = \rho(I)$, and can add a new triplet to a replay buffer: the program $\rho$, the demonstration inputs $I$, and the realized outputs $O$ (which may or may not match the target outputs $O^*$).
In each sampling stage we repeat this procedure $n_\rho$ times per task, where $n_\rho$ is a hyperparameter.

Replacing the target output by the realized one is a form of hindsight experience replay \citep{andrychowicz2017hindsight}, and ensures that we obtain an experience every time we find a syntactically correct program, thereby preventing stagnation of the buffer.
Although these programs may not solve the tasks we are interested in, they are always valid in terms of syntax and semantics (correctly mapping $\rho(I) \rightarrow O$).
They can therefore be used to teach the policy about program syntax and program behaviour, which may lead to positive transfer to the search set.
We emphasize that we never add test examples nor performance on the test examples to our buffer, as one should not have access to their target output grid during sampling.

\paragraph{Learning}

During the learning stage, the policy $Q_\theta$ is trained on experiences sampled from the buffer, the training set and the augmented training set. 
These experiences consist of input grids $I$, output grids $O$ and the corresponding program $\rho$.
The training objective is then a straightforward negative log-likelihood objective:
\begin{equation}
\mathcal{L}( \rho, I, O ) =  - \log Q_\theta( \rho | I, O ).
\end{equation}
We keep only a single copy of the policy network, and continue to update it during each learning stage. 
Since we do not compare the policy with its past versions there is no guarantee for improvement.
Although continual updates could lead to worse performance in the next iteration, we find this is not a problem in practice. 

By default, we perform prioritized sampling from the replay buffer \cite{schaul2015prioritized}. 
For each experience, the priority is proportional to the percentage of demonstration outputs equal to program outputs.
This means that programs that solve real ARC tasks' demonstration examples are sampled more often than programs for hindsight-relabeled tasks. 

\section{Experiments}

In this section, we aim to demonstrate the efficacy of CodeIt, and break down how much different components of the method contribute to the performance. 
We first tuned hyperparameters on a custom training and validation split (for a description of these parameters and details, see Appendix \ref{appendix:experiment details}).
Using these hyperparameters, we benchmark our method on the ARC evaluation split and compare against previous state-of-the-art methods.
Finally, we ablate the importance of individual components of CodeIt.

We define \emph{demonstration performance} as the percentage of solved demonstration examples on a given task. 
We first sort solution programs by demonstration performance, and then by program length, favoring shorter programs.
We evaluate the top three programs on the set of test examples.
Following ARC evaluation procedure, if at least one of these three programs maps all test example inputs to outputs, the task is solved and \emph{test performance} is 1.
We emphasize that the ExIt procedure only makes use of demonstration examples, and that we use test performance for final evaluation only.

\paragraph{Custom baselines}

We use a random baseline that samples programs line-by-line.
At the start of each line, we sample a primitive function from the DSL, then sample arguments given its expected input types. 
When a variable of type ``grid'' is created, we end the program with probability 0.8, otherwise we add another line to the program.

We also use a mutation-based baseline.
This is a more advanced procedure, designed with the DSL in mind.
At every meta-iteration, it mutates the set of training programs provided by \citet{hodel}.
We use two variations: ``$d_1$'' mutates only the initial training set, and ``$d_\infty$'' can augment newfound programs as well.
We provide the exact algorithm in Appendix~\ref{subsec:mutation}.

For all three baselines, we sample $n_m = n_\rho \cdot n_{tasks}$ programs per meta-iteration. 
Here, 
$n_\rho$ is the desired number of programs per meta-iteration per task, and
$n_{tasks}$ the total number of tasks in the population. 
To strengthen these baselines, we exhaustively evaluate each found program on all inputs in the search set, and check the outputs against ARC output grids.

\paragraph{Baselines from literature}

We include approaches from literature as baselines as well.
A direct comparison is sometimes difficult, as not all baselines apply their method to the full ARC evaluation set:
for example, ~\citet{kolev2020neural} and ~\citet{arc_neural_search} focus only on a subset of ARC.
Additionally, some symbolic methods design a DSL based on both ARC training and evaluation sets and report results on a hidden test set \cite{kaggle}.
We therefore only compare to approaches that report scores on the full ARC evaluation set.

\citet{arc_vision_image} and~\citet{ferre2023tackling} both run a search procedure for a custom DSL on the full set.
As~\citet{arc_vision_image} report the highest performance the full ARC evaluation set, this is our main symbolic baseline.
Although \citet{mirchandani2023large} and \citet{gendron2023large} use a different evaluation protocol, we include these as our main neural baseline, as they are based on powerful LLMs (text-davinci and GPT-4).

\subsection{Setup}

We initialize our training set with the 400 examples from the ARC training split and the associated solution programs provided by \citet{hodel}. 
We also sample 19,200 programs as additional training data via the mutation procedure outlined in Appendix \ref{subsec:mutation}.
We use the programs that are syntactically correct to initialize the augmented training set. 
We use the 400 ARC evaluation examples as our search set.

In the sampling stage of each meta-iteration, we use temperature sampling with temperature $\tau=0.95$, and sample up to $n_\rho=24$ programs per task.
This encourages exploration and, as a result, increases the diversity of data added to the replay buffer. 
We reject policy-sampled programs if they are syntactically incorrect, or if they run for more than 0.25 seconds per program line.
All valid programs are added to the replay buffer.
 
 In each learning stage, we start by sampling a set of experiences from the buffer under the distribution given by the priorities. Each meta-iteration, we sample $r_t=10,000$ experiences from the concatenation of the train set and the augmented train set, and $r_p=90,000$ experiences from the buffer. The resulting set is used for 1 epoch of training.
For a full list of hyperparameters, see Table \ref{tab:hyperparameters} in the Appendix.

\begin{figure}
    \centering
    \includegraphics[width=0.49\textwidth]{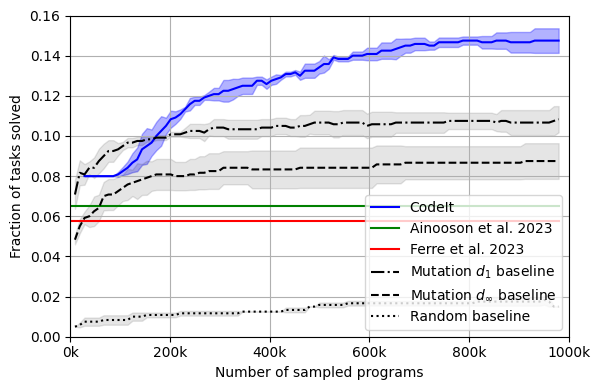}
    \caption{Cumulative performance as function of number of sampled programs for CodeIt and various baselines, showing mean and standard deviation of three runs for CodeIt and custom baselines.}
    \label{fig:cumulative_performance_vs_baselines}
\end{figure}


\begin{table*}[t]
    \centering
    \begin{tabular}{l r r r}
    \toprule
        Method                                 & ARC Train Set & ARC Eval Set & ARC Eval 412\\
     \midrule                                  
        \citet{ferre2021first}                 & 29 / 400      & 6 / 400   & - \\
        \citet{arc_vision_image} MLE           & 70 / 400      & 17 / 400  & - \\
        \citet{arc_vision_image} brute force   & 104 / 400     & 26 / 400  & - \\
        \citet{ferre2023tackling}              & 96 / 400      & 23 / 400  & - \\
        \citet{mirchandani2023large} text-davinci-003
           & 56 / 400\textsuperscript{*}  
           & 27 / 400\textsuperscript{*} & - \\
        \citet{gendron2023large} GPT-4 & - & - & 49 / 412\textsuperscript{*} \\
        \midrule 
        Mutation $d_1$ baseline                 & -       & 42 / 400   &  39 / 412\textsuperscript{*}     \\
        Mutation $d_\infty$ baseline            & -       & 38 / 400   &  36 / 412\textsuperscript{*}     \\
        Random baseline                         & -       &  6 / 400   &   7 / 412\textsuperscript{*}    \\
        \midrule                              
        CodeIt               & -           & \textbf{59 / 400}  & \textbf{59 / 412\textsuperscript{*}}   \\
    \bottomrule
    \end{tabular}
    \caption{Main results on ARC eval set. The evaluation metric is pass@3 by default, \textsuperscript{*} indicates pass@1. To enable comparison to related work of \citet{gendron2023large}, we also include pass@1 performance on the ARC Eval set with 412 examples. Our method outperforms all previous baselines. More details on the ARC splits and evaluation procedures can be found in Appendix \ref{appendix:evaluation}.}
    \label{tab:results}
\end{table*}

\subsection{Main results on ARC eval set}

In Figure~\ref{fig:cumulative_performance_vs_baselines}, we show performance as a function of the number of sampled programs, for CodeIt, our custom baselines, \citet{arc_vision_image} and \citet{ferre2023tackling}.
We show \emph{cumulative performance} here, which means that any program in the buffer or augmented train set is considered a solution candidate.
For the mutation baselines, we see a rapid performance increase followed by stagnation. 
In comparison, CodeIt takes several meta-iterations to start generating solutions outside of the augmented train set and then performance rapidly increases.
CodeIt quickly outperforms the mutation baseline, indicating that it indeed finds higher-quality samples to train on.

We report final performance of CodeIt after 100 meta-iterations, and the performance of various baselines, in Table~\ref{tab:results}.
To enable comparison to \citet{gendron2023large}, we include results on the ``ARC Eval 412'' set, which treats each test example in the ARC evaluation set as a separate task.
Our approach outperforms symbolic approaches ~\cite{arc_vision_image, ferre2021first, ferre2023tackling}, but also neural approaches based on large language models \cite{gendron2023large, mirchandani2023large},
achieving state-of-the-art performance on the ARC evaluation set.

\begin{figure*}
    \centering
    \includegraphics[width=1\textwidth]{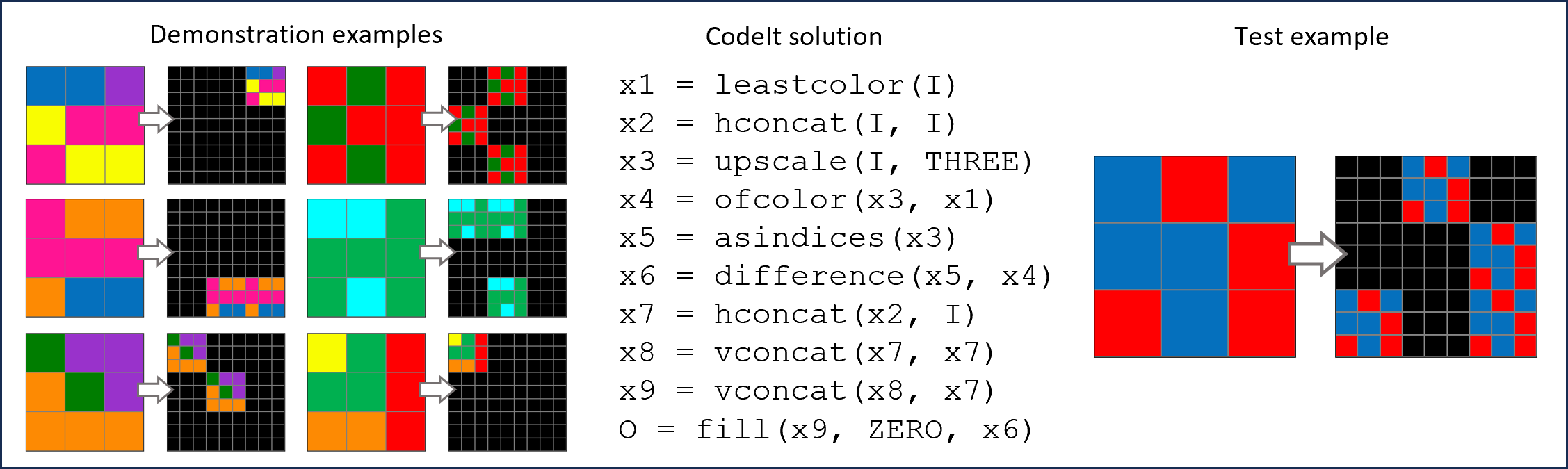}
    \caption{ARC evaluation task 48f8583b and the solution program found by CodeIt.}
    \label{fig:example_solution}
\end{figure*}

For context, we show a solution written by CodeIt for an example task in Figure \ref{fig:example_solution}.
To further illustrate the differences between the programs found by CodeIt and the mutation baselines, we analyze solutions found by each method in Appendix \ref{appendix:codeitvsmutation}, including a qualitative comparison in Table \ref{tab: task_solved_per_method}. 
One finding is that there are 29 tasks for which CodeIt and the mutation baseline both find a solution, but that there are 23 tasks for which only CodeIt finds a solution, versus 13 for the mutation baseline. 
For the tasks that both methods solve, CodeIt finds shorter programs on average and uses different primitives. 

In Appendix \ref{appendix:codeitvstime},
we observe CodeIt refines its initial solution for 53\% of solved tasks, producing a shorter solution in a later meta-iteration.
 Moreover, in Appendix \ref{subsec:failure_cases}, we analyze failure cases on the customer validation set and observe that CodeIt does not solve tasks with solution programs longer than 11 lines but often learns to use their primitives. Further, in Appendix \ref{subsec:primitives}, we look at the DSL primitives that CodeIt learns and find that some primitive types are learned quicker than others. Finally, in Appendix \ref{subsec:concept_arc}, we find that CodeIt appears to perform best on tasks related to object interactions and worst on numerical or logic based tasks by analyzing performance on the ConceptARC dataset \citep{moskvichev2023conceptarc}.

\subsection{Ablations}

\begin{figure}
    \centering
    \includegraphics[width=0.49\textwidth]{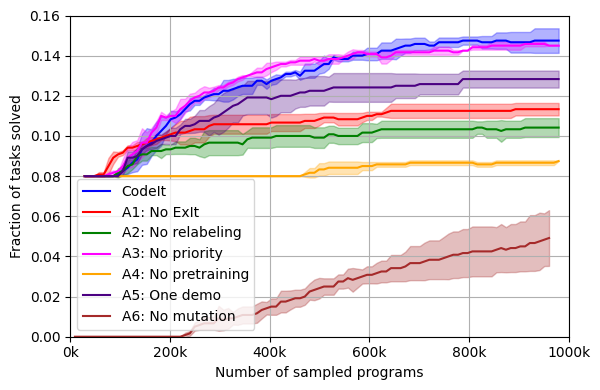}
    \caption{Cumulative performance as function of number of sampled programs for CodeIt and ablations, for three random seeds. For cumulative performance, all programs in the augmented train set and buffer are candidate solutions.}
    \label{fig:cumulative_performance_vs_ablations}
\end{figure}

\begin{figure}
    \centering
    \includegraphics[width=0.49\textwidth]{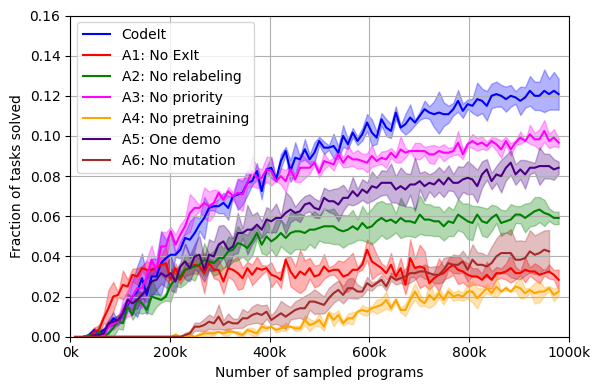}
    \caption{Policy performance per meta iteration as function of number of sampled programs for CodeIt and ablations, for three random seeds. For policy performance, only programs output by the policy in the \emph{current} meta-iteration are candidate solutions.}
    \label{fig:performance_vs_ablations}
\end{figure}

In Figure \ref{fig:cumulative_performance_vs_ablations} and \ref{fig:performance_vs_ablations}, we report cumulative performance and policy performance over time for CodeIt and all ablations. 
In all cases, we initialize the method with the ARC train set, and use the ARC evaluation set as search set.
We show the results of ablations at the end of training in Table \ref{tab:ablations}.
We also perform a scaling study with different model sizes, results are shown in Appendix \ref{sec:scaling_study}.

\paragraph{A1: No ExIt} 
This ablation removes policy feedback, to isolate the contribution of Expert Iteration.
In every meta-iteration, instead of populating the buffer with policy samples, we take the programs generated in that meta-iteration of the mutation $d_1$ baseline. 
For each program, we randomly select a task from the search set and perform hindsight relabelling, adding the program, input, output triplet to the buffer. 
We sample $r_p+r_t=100,000$ experiences from the concatenation of the train set, the augmented train set and the buffer at each meta-iteration for learning. 
We see that A1 outperforms the mutation baseline, which means supervised learning from mutation experiences alone does lead to some inter-task generalization.
However, cumulative performance is substantially lower than CodeIt. 
This highlights the importance of policy feedback.

\paragraph{A2: No relabeling} 
We test the effect of hindsight relabeling by only adding experiences to the buffer if the program produces the correct output for all demonstration examples.
We train on all experiences in the buffer without prioritized sampling.  
Although performance increases in early meta-iterations, A2 stagnates after around 30 meta-iterations, indicating that data generated by sampling and filtering alone is not sufficient. 
Sampling and hindsight relabeling (CodeIt) performs better than sampling and filtering (A2).

\paragraph{A3: No priority} 
To test the hypothesis that prioritized sampling ameliorates catastrophic forgetting, we draw experiences uniformly from the buffer in the learning stage. 
A3 leads to a small reduction in cumulative performance, but a large reduction in policy performance, indicating that the policy indeed forgets important experiences. 
Prioritized sampling results in better retention of knowledge.

\paragraph{A4: No pretraining} 
To identify whether our pre-trained policy contains beneficial prior knowledge, we randomly reinitialize the policy’s weights at the start of CodeIt. 
Policy performance shows that performance improvement is much slower.
Moreover, inter-task generalization begins later, as shown by the cumulative performance, which only starts increasing after around 50 meta-iterations.
Despite the expected slowdown, it is encouraging to see that CodeIt does seem to be able to bootstrap from random weights.

\paragraph{A5: One demo} We investigate 
use of the task representation by decreasing the number of demonstration examples shown to the policy. 
This results in a strong decrease in both cumulative and policy performance. 
This indicates CodeIt forms abstractions over multiple demonstration examples.

\paragraph{A6: No mutation} 
In this ablation, we omit the mutation-based training data augmentation step.
We observe that taking out mutation-based bootstrapping results in slower training, although performance does increase over time and does not stagnate.
We therefore conjecture that mutation-based augmentation is not strictly required but still useful.


\begin{table*}
    \centering
    \begin{tabular}{l r r r r r}
    \toprule
                                  & initial policy & \# demo      &  \# policy  &policy only&  cumulative               \\
        Method                    & weights &  examples     &  samples &   perf.&perf.        \\
     \midrule                                 
        CodeIt                    & CodeT5     & $\leq 10$      & 24       &   49/400   &  59/400          \\
     \midrule                                        
        A1: No ExIt    & CodeT5     & $\leq 10$      & 0       &   13/400     & 45/400           \\
        A2: No relabeling    & CodeT5     & $\leq 10$      & 24      &      24/400   & 42/400           \\
        A3: No priority & CodeT5     & $\leq 10$      & 24      &    38/400   &  58/400          \\
        A4: No pretraining & Random     & $\leq 10$      & 24     &   9/400     &  35/400          \\
        A5: One demo & CodeT5     & $\leq 1$      & 24      &   34/400    &  51/400          \\
        A6: No mutation & CodeT5     & $\leq 10$      & 24      &   17/400   &  20/400          \\
    \bottomrule
    \end{tabular}
    \caption{ARC evaluation performance of CodeIt ablations.}
    \label{tab:ablations}
\end{table*}

\section{Related work}
\subsection{Abstraction and Reasoning Corpus (ARC)}

Various works have applied program synthesis approaches to subsets of the ARC dataset.
\citet{graphsarc} proposes to represent grids as graphs, and applies logical programs to the graph nodes, solving 63 of 160 tasks.
\citet{kolev2020neural} apply a Differentiable Neural Computer to ARC, solving 78\% of tasks with grids of size $10\times10$ and smaller.
\citet{alford2022neuralguided} applies DreamCoder \citep{ellis2020dreamcoder} and execution-guided program synthesis, solving 22 of 36 considered tasks.
\citet{park2023unraveling} first collects human feedback, then performs behavioral cloning for a subset of ARC tasks using a decision transformer \citep{decision_transformer}.
However, none of these methods are applied on the full ARC evaluation set, typically due to poor scaling behavior. 

The few works that do scale to the full evaluation set tend to solve each task in isolation.
\citet{ferre2021first} and followup work \citet{ferre2023tackling} design a custom DSL and perform a fast search for each task.
\citet{arc_vision_image} also designs a custom DSL and obtains best performance with a brute-force search, solving 36 of 400 evaluation tasks.
\citet{mirchandani2023large} and \citet{gendron2023large} demonstrate that a pretrained language model with custom tokenizer will output the correct grid after being shown multiple input-output pairs, solving 27 of 400 and 49 of 412 evaluation tasks respectively.
\citet{wang2023hypothesis} augment this approach by generating hypotheses in multiple rounds, although they only show performance on a subset of the ARC training set due to the high monetary cost of querying the language model.
In this work, we design a scalable ExIt approach that combines a smaller language model with the higher-level abstraction of a DSL.
We also ensure that our approach incorporates experience to benefit from generalization between tasks.

Various unpublished approaches exist too, including submissions to ARC challenges as well as a Kaggle competition.
These competitions use a private leaderboard, not revealed to participants.
This means participants often use the public ARC evaluation set for training or DSL design purposes.
For example, the winner of Kaggle 2020 comments that searching in a DSL designed using the training set resulted in low performance, and higher performance was reached after conditioning the DSL on the evaluation tasks \cite{kaggle}. 
This makes direct comparisons to methods evaluated on the evaluation set difficult. 
For reference, we include a summary of competition results in Appendix \ref{sec:arc_competitions} Table \ref{tab:arc_competitions}, however, note that this summary reports performance on the hidden test set, and that competition results cannot be directly compared to this work and the literature.


\subsection{Expert Iteration}
Expert iteration (ExIt) \citep{Anthony2017-hb} consists of a policy-guided search stage that gathers new experiences, and a learning stage that improves the policy by imitation learning. 
Commonly used experts tend to be powerful and computationally intensive tree search algorithms such as Monte Carlo Tree Search~\citep{kocsis2006bandit} and greedy search~\citep{daume2009search}. 
ExIt has achieved superhuman performance include games \citep{silver2016mastering, silver2018general, Anthony2017-hb} and combinatorial problems such as bin-packing \citep{laterre2019ranked}. 
Related work that employ hindsight relabeling in expert iteration are \citet{aygün2021proving}, \citet{butt2022program} and \citet{gauthier2022learning}.

Applications of ExIt for programming-by-examples \cite{mankowitz2023faster, ellis2020dreamcoder} are most relevant to CodeIt. 
\citet{mankowitz2023faster} consider one task only: writing a fast sorting algorithm. For this problem, inter-task generalization is thus not as important. 
DreamCoder \cite{ellis2020dreamcoder} is most related to our work, since this ExIt method is applied to multiple programming-by-examples tasks. 
DreamCoder uses a continually growing DSL to store abstractions, and a computationally intensive search procedure. Instead, CodeIt uses the model to store distilled knowledge, and generates experiences via sampling from the model. 
Furthermore, DreamCoder filters solutions based on correctness whereas CodeIt uses hindsight relabeling and prioritized experience replay. 

\subsection{Self Improving Large Language Models}

Previous work showed that learning from synthetic data is a viable strategy for theorem proving \cite{wang2020proving} and programming-by-examples \cite{balog2017deepcoder, devlin2017robustfill, bunel2018leveraging,  parisotto2017neuro, polosukhin2018neural, zohar2018automatic}, often training a model from scratch. 
Instead, finetuning pre-trained large language models (LLMs) on synthetic data enables knowledge transfer due to the prior domain knowledge captured in their weights 
\cite{butt2022program}.
Recently, methods that use LLMs to synthesize training data have shown successes in general domains including 
theorem proving \cite{polu2022formal}, 
question answering \cite{zelikman2022star, aksitov2023rest}, 
mathematical reasoning \cite{ni2023learning}, 
machine translation \cite{gulcehre2023reinforced}, 
language-to-code generation \cite{zhou2023language, singh2023human} and code-to-code generation \cite{haluptzok2022language}. 
We demonstrate in this work that such an approach can be applied to the challenging ARC domain as well.

\section{Discussion}


Various factors make ARC uniquely challenging for learning-based approaches, for example the limited amount of training data, and the complexity of individual tasks.
Another issue is that programs may differ in number of demonstration examples and input dimensionality, which requires agents to reason about concepts at different scales.  
In this work, we show that an expert iteration based approach can learn to solve 59/400 unseen ARC tasks.
Here, we provide intuition for why CodeIt works well on this benchmark.


Ablations showed that hindsight relabeling has a large effect on performance.     
Many expert iteration approaches rely on the emergence of a curriculum of increasingly difficult tasks, even creating a  curriculum by comparing the current agent to past versions of itself \cite{silver2016mastering, alphatensor} or reward shaping \cite{laterre2019ranked, gulcehre2023reinforced}.
Hindsight relabeling forms an implicit curriculum \cite{andrychowicz2017hindsight}: initially we collect easy tasks that can be solved in few lines of code, while later on, programs become more complex.
This is useful for ARC, where obtaining even one solved task is challenging.
As relabeling adds many programs to the buffer, including some that are further away from the target tasks, we used prioritized sampling to avoid catastrophic forgetting.



A potential limitation of CodeIt is that for ARC, it relies on hand-designed components: a domain specific language (DSL), access to an interpreter for automatic evaluation, and an initial set of ground truth programs. 
While we do benefit from Hodels expert-designed DSL, we also showed that a neuro-symbolic approach (ablation A1) outperforms a symbolic approach (the mutation baseline), indicating that both DSL and learning contribute to performance. 
Further, CodeIt outperforms both, indicating that ExIt compounds this effect.
We also use a pretrained LLM and mutation procedure to speed up training, but ablations showed that training is possible even without these, albeit at a slower pace.
Nevertheless, approaches that can start learning tabula rasa, or form their own DSL \cite{ellis2020dreamcoder} remain an important area of research.



For the ARC dataset, it is currently beneficial to incorporate both prior knowledge (via a DSL or pre-trained LLM) and experience (via expert iteration).
\citet{chollet2019measure} 
defines the intelligence of a system as ``a measure of its skill-acquisition efficiency over a scope
of tasks, with respect to priors, experience, and generalization difficulty''. 
Chollet poses that, if two systems are initialized with the same prior knowledge and go through the same amount of experience with respect to a set of unseen tasks, the more intelligent system will combine prior knowledge and its experience more efficiently, solving more tasks. 

Although many existing approaches incorporate prior knowledge through a programming language or DSL \cite{arc_vision_image,ferre2023tackling}, a pre-trained large language model \cite{gendron2023large, mirchandani2023large}, or both \cite{wang2023hypothesis}, they cannot incorporate new experience, and therefore do not benefit from inter-task generalization. 
\citet{alford2021neurosymbolic} proposes an expert iteration method that does learn from experience, but it does not scale well nor benefit from prior knowledge in its policy. 
We pose that CodeIt is the more effective expert iteration method due to its use of scalable components: pre-trained language models, likelihood-based training, and running programs in interpreters.
There is also an implicit relationship between \emph{computational} efficiency and experience: since CodeIt's policy learns on the ARC domain, it is possible to use a much smaller language model than for example \citet{gendron2023large}, who use GPT-4 as a policy. 
This is consistent with LLM literature showing that high quality training data with a curriculum enables smaller LMs to compete with much larger ones on coding tasks \cite{gunasekar2023textbooks}.

\section{Conclusion}

We introduce a novel and scalable method for self-improving language models, \emph{CodeIt}, that uses prioritized hindsight replay. 
CodeIt achieves state-of-the-art performance on the Abstraction and Reasoning Corpus (ARC) compared to symbolic and neural baselines, solving 59 of 400 evaluation tasks.
Ablations show that hindsight relabeling leads to improved sample efficiency resulting in a 40\% improvement in performance. 
We also find that prioritizing important experiences during training ameliorates catastrophic forgetting. 
Additionally, we observe that CodeIt is able to refine solutions over time, identifying a shorter program for 53\% of solved tasks in later iterations. 
The results demonstrate that our self-improving language model is capable of reasoning in the program space and generalizing between tasks.
 For the challenging ARC benchmark, both scalability and learning from experience prove to be key components for success.

\section*{Acknowledgements}
We thank Michael Hodel for the creation of their DSL for solving ARC tasks, available at \url{https://github.com/michaelhodel/arc-dsl}. 

\section*{Impact Statement}

This work presents a method for training a language model based policy on programming-by-example problems where data is scarce.
Our approach learns to write programs in iterations, and receives feedback from an interpreter.

On one hand, large language models have potential negative societal impacts, exacerbating biases present in training data.
This can be especially harmful if datasets are small, as is the case in our work.
Additionally, models that generate code can be used to automate writing of malicious code.

On the other hand, enabling the use of large language models on problems where data is scarce opens up application areas where it is currently challenging to deploy them.
Models that generate code from input-output specifications are broadly applicable and can be used in many programming settings.

\bibliography{refs}
\bibliographystyle{icml2024}

\newpage
\appendix
\onecolumn
\clearpage
\appendix
\section{Method and evaluation details}

\subsection{CodeIt Algorithm}
\label{subsec: pseudo}

The pseudo code for the CodeIt procedure is portrayed in Algorithm~\ref{alg:codeit}. 

\begin{algorithm}
\caption{CodeIt Algorithm}
\label{alg:codeit}
\begin{algorithmic}
    \REQUIRE Training set \(D_{\text{train}}\), search set \(D_{\text{test}}\), policy \(Q\)
    \ENSURE Finetuned policy \(Q\), updated replay buffer \(R\), optimal programs set \(\rho^*\)

    \STATE \(D_{\text{augmented\_train}} \gets \text{EvolveTrainingTasks}(D_{\text{train}})\) \COMMENT{Evolve training tasks}
    \STATE \textbf{Initialize} \(\rho^*\) as an empty set \COMMENT{Init set of programs that solve tasks in $D_{test}$}
    \FOR{\(\text{meta\_iter} = 1 \to 100\)} 
        \STATE \textcolor{blue}{\# Sampling and hindsight relabeling stage}
        \FOR{\text{task} in \(D_{\text{test}}\)}
            \STATE \(\{\rho\} \gets Q(\rho|\{I, O\})\) \COMMENT{Sample programs for test tasks}
            \FOR{each \(\rho\) in \(\{\rho\}\)}
                \IF{\(\text{SyntacticallyValid}(\rho)\)}
                    \STATE Add \(\{\rho,\{(I^{(i)}, \rho(I^{(i)} )), \dots \}\}\) to \(R\) \COMMENT{Update the replay buffer with hindsight relabeled tasks}
                \ENDIF
                \FOR{\((I^{(i)}, O^{(i)})\) in task}
                    \IF{\(\rho(I^{(i)}) = O^{(i)}\)}
                        \STATE Add \{(\(\rho\), \text{task})\} to \(\rho^*\) \COMMENT{Update set of programs that solve tasks in $D_{test}$}
                    \ENDIF
                \ENDFOR
            \ENDFOR
        \ENDFOR
        
        \STATE \textcolor{blue}{\# Learning stage}
        \STATE \(D_{\text{sample}} \gets \text{SampleFrom}(R+D_{\text{augmented\_train}}+D_{\text{train}})\) \COMMENT{Sample tasks from the replay buffer}
        \STATE Train \(Q\) on \(D_{\text{sample}}\) for 1 epoch \COMMENT{Continual training of the policy}
    \ENDFOR

\end{algorithmic}
\end{algorithm}

\paragraph{Initializing CodeIt} 
\label{appendix:initmutation}

Before we start the CodeIt procedure, we expand the training dataset using the first 19,200 mutated tasks from the mutation procedure (see Appendix \ref{subsec:mutation}) used for the mutation $d_1$ baseline.

\subsection{Program and Task Mutation}
\label{subsec:mutation}

\paragraph{Mutation procedure} 

To grow a population of mutated programs with task demonstration inputs corresponding to the original training dataset, we follow the procedure outlined in Algorithm \ref{alg:evolve_training_tasks}. This involves mutating a single task, which is described in Algorithm \ref{alg: mutate}. The mutation is carried out with the hyperparameters $\phi_{\text{var}}=0.25, \phi_{\text{arg}}=0.5, \phi_{\text{func}}=0.25$. With respect to naming notation, $d_1$ reflects a depth of 1, meaning we only mutate programs from the original training set, and $d_\infty$ reflects a depth of infinity, meaning we can mutate previously mutated programs.

The intuitive explanation of the mutation procedure for a single program is as follows.
We pick a random line from a program (L2-3).
We then replace either a function call with a function with similar output type (L4-7), or we replace an input argument in the function call (L8-11), or we replace the function call but leave its input variables the same (L12-14).

\begin{algorithm}
\caption{MutateProgram}
\label{alg: mutate}
\begin{algorithmic}
\REQUIRE{Replacement probabilities $\phi_{\text{var}}$, $\phi_{\text{arg}}$, $\phi_{\text{func}}$, program $\rho$} 
\ENSURE{$\rho'$} 
    \STATE \textbf{Initialize} $\rho' \gets \rho$ \COMMENT{Copy original program}
    \STATE $l \gets \text{RandomLineFrom}(\rho')$ \COMMENT{Randomly select a line}
    
    \STATE $p \sim U(0, 1)$ 
    
    \IF{$p < \phi_{\text{var}}$} 
        \STATE $f' \gets \text{SampleFunctionWithOutputType}(\text{GetTypeOfVariable}(l))$
        \STATE $args' \gets \text{SampleArgumentsForFunction}(f')$
        \STATE Replace variable definition $f(args)$ in $l$ with $f'(args')$
    \ELSIF{$p < (\phi_{\text{var}} + \phi_{\text{arg}})$} 
        \STATE $a \gets \text{RandomArgumentFrom}(l)$ 
        \STATE $a' \gets \text{SampleTermOfType}(\text{GetTypeOfArgument}(a))$
        \STATE Replace argument $a$ with $a'$
    \ELSE 
        \STATE $f' \gets \text{SampleFunctionOfType}(\text{GetTypeOfFunction}(f))$
        \STATE Replace function $f$ in $l$ with $f'$
    \ENDIF
    
\end{algorithmic}
\end{algorithm}

\begin{algorithm}
\caption{EvolveTrainingTasks}
\label{alg:evolve_training_tasks}
\begin{algorithmic}
    \REQUIRE{Initial population of training tasks \( T_{\text{init}} \) (each task is a tuple \( (\rho, \mathcal{E}) \) where \( \mathcal{E} = \{(I^{(i)}, O^{(i)}), \dots\} \)), depth}
    \ENSURE{Updated task population \( T' \) (initialized with \( T_{\text{init}} \))}

    \STATE \( T \gets T_{\text{init}} \) 
    \STATE \( i \gets 0 \)

    \WHILE{\( i < num\_samples \)}
        \IF{\( \text{depth} = 1 \)}
            \STATE \( (\rho, \mathcal{E}) \gets \text{RandomSelectTask}(T_{\text{init}}) \) \COMMENT{Select from initial tasks}
        \ELSE
            \STATE \( (\rho, \mathcal{E}) \gets \text{RandomSelectTask}(T) \) \COMMENT{Select from current tasks}
        \ENDIF
    
        \STATE \( \rho' \gets \text{MutateProgram}(\rho) \) 
    
        \STATE \( \mathcal{E}' \gets \emptyset \) \COMMENT{Initialize mutated task demonstration examples}
        \FOR{each \( (I^{(k)}, \_) \in \mathcal{E} \)}
            \STATE \( O'^{(k)} \gets \text{Execute}(\rho', I^{(k)}) \)
            \STATE \( \mathcal{E}' \gets \mathcal{E}' \cup \{(I^{(k)}, O'^{(k)})\} \)
        \ENDFOR

        \IF{\( \text{AreValidGrids}(\text{GetAllOutputs}(\mathcal{E}')) \)}
            \STATE \( T' \gets T' \cup \{(\rho', \mathcal{E}')\} \) \COMMENT{Add new task to the population}
        \ENDIF

        \STATE \( i \gets i + 1 \)
    \ENDWHILE

\end{algorithmic}
\end{algorithm}

\paragraph{Mutation baseline} 
For our mutation baseline, we sample mutated programs using the mutation procedure outlined above. For all the mutated programs in the evolved task population, we evaluate each program on the tasks in our search set.

\subsection{Task Representation}
\label{subsec:task_representation}

\paragraph{Grid representation}

We use a compressed grid representation, mainly to reduce the number of tokens needed to represent each grid.
We do not use a custom tokenizer.
A visualization of the number of tokens is shown in Fig.~\ref{fig: grid_tokens}, showing that in almost all cases, the sparse grid representation we use leads to a reduction in the number of needed tokens, especially for larger grid sizes.

\paragraph{Truncation} 
We truncate our task demonstration tokens and program tokens such that these sequences fit in our predefined encoder and decoder context windows. For the task demonstration examples, we first order by grid size and divide the encoder context window into two equally sized sections. For task demonstration inputs, we first encode input grids to text as above and then we tokenize using the standard text tokenizer. We truncate these tokens at half the size of the encoder context window. We do the same for the task demonstration outputs and with the exception of also adding an end of sequence token. As a result, even though we aim to show the policy up to ten task demonstration examples, large grids will be cut-off. For programs, we tokenize directly using the standard text tokenizer and truncate at the decoder context window size.

\subsection{ARC evaluation}
\label{appendix:evaluation}

Different works use different evaluation procedures to report performance on the ARC evaluation set.   
We describe two common evaluation settings in more detail below. 
Unless mentioned otherwise, we always use the first procedure, ``ARC Eval Set''.

\paragraph{ARC Eval Set}
This setup is intended as close as possible to the evaluation procedure described by \citet{chollet2019measure}. 
Baselines \citet{ferre2021first}, \citet{arc_vision_image} follow this procedure, and it is our default setting as well.

The ARC eval set consists of 400 tasks, some of which contain multiple test examples.
Common procedure is to report pass@3 performance, meaning the top 3 solutions are selected according to demonstration task performance.
If there are ties, we favor the shorter program, under the assumption that shorter programs are more likely to generalize.
We then run these programs on all test examples for the task.
In some cases, there are multiple test examples per task.
We call the task ``solved'' if all output grids are correct.

\paragraph{ARC Eval 412}
This setup is designed to match \citet{gendron2023large}.
Instead of calling a task with multiple test examples solved if all test outputs are correct, distinct tasks are created - one per test example.
This results in a set of 412 evaluation tasks with one test example each.
Furthermore, \citet{gendron2023large} uses pass@1, rather than pass@3: only one solution per task is evaluated, and the task is considered solved if the output is correct.

\section{Experiment details} 
\label{appendix:experiment details}
\subsection{Resources}
Experiments were run for a maximum of 120 hours on a NVIDIA A100 80GB.

\subsection{Hyperparameter tuning}

\paragraph{Dataset}
The ARC benchmark does not contain a validation split.
Hence, we use part of the ARC train split for validation during the hyperparameter tuning. In particular, this validation set is the search set that the sampling stage uses as described in \ref{ssec:sampling stage}. With this setup we avoid overfitting the hyperparameters to the ARC evaluation split. 

We choose the split such that $\train$ and $\valid$ contain roughly equally difficult programs by sampling based on program length: $\train$ contains 80\% of 2-line programs, 80\% of 3-line programs, and so on.
This results in 311 examples in $\train$ and 89 examples in $\valid$.

\paragraph{Experiments on validation set}
In these experiments, we initialise our replay buffer with the 311 $\train$ examples, and our search set consists of the 89 $\valid$ examples. 
The aim of these experiments is to find optimal hyper-parameters for search and training. 
A list of our tuned hyperparameter values and their description is shown in Tab.~\ref{tab:hyperparameters}

\subsection{Hyperparamaters chosen on internal validation set}

We optimized these parameters on our custom validation set before applying CodeIt to ARC eval.

\label{subsec:hyper_params}
\begin{table}[H]
    \centering
    \begin{tabular}{c c c l}
    \toprule
    \textbf{CodeIt stage} & \textbf{Param} & \textbf{Value} & \textbf{Description} \\
    \midrule
    \multirow{2}{*}{Sampling and Hindsight Relabeling} 
    & $n_\rho$ & $24$ & no. policy samples $\rho$ per task per meta-iteration\footnotemark[1] \\
    & $n_m$ & $19,200$ & no. mutated samples for augmented train set\footnotemark[1] \\
    & $\tau$ & $0.95$ & sampling temperature \\
    & $r_t$ & $10,000$ & number of experiences sampled from augmented train set \\
    & $r_p$ & $90,000$ & number of experiences sampled from buffer \\
    \midrule
    \multirow{3}{*}{Learning} 
    & $n_\epsilon$& $1$ & no. train epochs per meta-iteration \\
    & $lr$& $5e-5$ & learning rate \\
    \bottomrule
    \end{tabular}
    \caption{Table of hyperparameters.}
    \label{tab:hyperparameters}
\end{table}
\footnotetext[1]{Note that no. samples here refers to policy and mutation samples before filtering for syntactic correctness.}

\subsection{Domain Specific Language}
\label{appendix:hodeldsl}
We adopt the domain specific language (DSL) of Michael Hodel, made available on GitHub: \href{https://github.com/michaelhodel/arc-dsl}{https://github.com/michaelhodel/arc-dsl}.
This DSL was designed based on the training set: the (human) designer did not peek at the evaluation set.
This is what allows us to run search on ARC eval here. 
Using a DSL designed for the eval tasks would be cheating, as we would benefit immensely from human insights captured in the primitives.
On the other hand, it may mean that some ARC eval programs are not solvable with the current DSL.

The DSL is implemented in \href{https://github.com/michaelhodel/arc-dsl/blob/main/dsl.py}{https://github.com/michaelhodel/arc-dsl/blob/main/dsl.py}.
It contains many basic grid manipulation operations, such as rotations (\texttt{rot90, rot180, rot270}), mirroring (\texttt{dmirror}, \texttt{hmirror}, \texttt{vmirror}), resizing (\texttt{downscale}, \texttt{upscale}), or concatenation (\texttt{hconcat}, \texttt{vconcat}).
It also contains functions that perform counting, for example \texttt{numcolors} counts the number of colors occurring in an object or grid.
For some ARC tasks, identifying the foreground objects and determining how these objects interact is an effective strategy for human test-takers.
Therefore, some functions also apply to ``objects'', which are patches of the same color that stand out from the background. 
To extract these, the function \texttt{objects} returns the set of foreground objects, i.e. those that have a different color than the most common color, assumed to be the background.
For a complete list of primitives and their description, we refer the reader to the aforementioned Github page.

Michael Hodel provides hand-designed solution programs for all training tasks in 
\href{https://github.com/michaelhodel/arc-dsl/blob/main/solvers.py}{https://github.com/michaelhodel/arc-dsl/blob/main/solvers.py}.
Some programs are highly complex: for some of the more challenging ARC tasks, we see solutions consisting of up to 58 lines of code (\texttt{solve\_b775ac94}).
We use these 400 solution programs to kickstart CodeIt training.
\section{ConceptARC}
\label{subsec:concept_arc}
The ConceptARC dataset \cite{moskvichev2023conceptarc} is similar to ARC with respect to the underlying task being a grid transformation problem, and the need to few-shot generalize from demonstration to test examples. However, tasks are divided into 16 concept groups; ConceptARC is moreover designed to be easier than ARC. We perform analysis on this additional benchmark, to highlight the type of tasks that CodeIt works well on.

We apply the same hyper-parameters as reported in Section \ref{subsec:hyper_params}, with the exception of $n_p=60$ which we rescale in proportion to the number of evaluation tasks in the ConceptARC dataset. The evaluation metric is also different: for ConceptARC, performance is evaluated separately on each test case per task. To ensure our results are comparable with existing work, we evaluate CodeIt by taking the model checkpoint at 100 meta-iterations, and perform pass@3 inference for temperature=0 (greedy decoding) and temperature=1 (simple sampling). 

\begin{figure}
    \centering
    \includegraphics[scale=0.55]{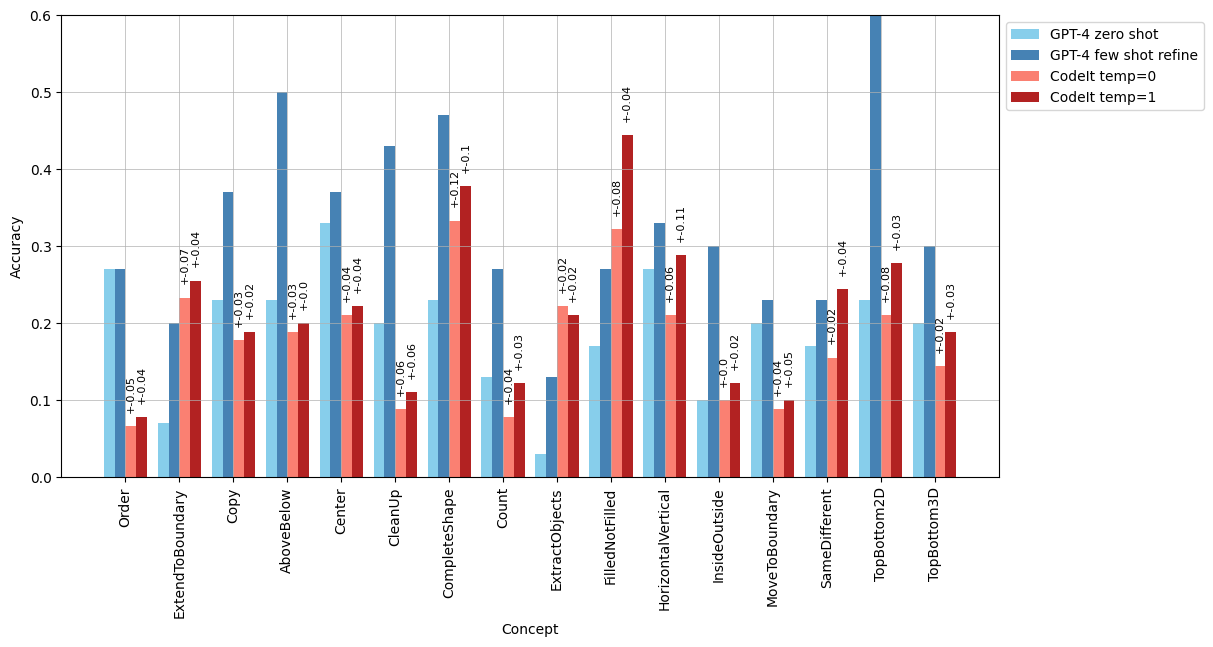}
    \caption{ConceptARC test accuracy CodeIt (three seeds) and GPT-4}
    \label{fig:conceptARC}
\end{figure}

In Figure \ref{fig:conceptARC}, we report CodeIt performance on ConceptARC with two evaluation temperatures against GPT-4 baselines from the literature, GPT-4 zero shot \cite{moskvichev2023conceptarc} and GPT few shot refine \cite{mitchell2023comparing}. 
We observe that CodeIt outperforms GPT-4 with respect to the following concepts: extend to boundary, extract objects, filled not filled and same different. However, CodeIt substantially under-performs GPT-4 with respect to order, clean up and count. We note that while the DSL includes primitives for ordering and counting, it appears that CodeIt performs best on tasks related to object interactions, however, CodeIt performs worst on numerical or logic based tasks.

\section{Scaling Study}
\label{sec:scaling_study}
\begin{figure}
    \centering
    \includegraphics[scale=0.7]{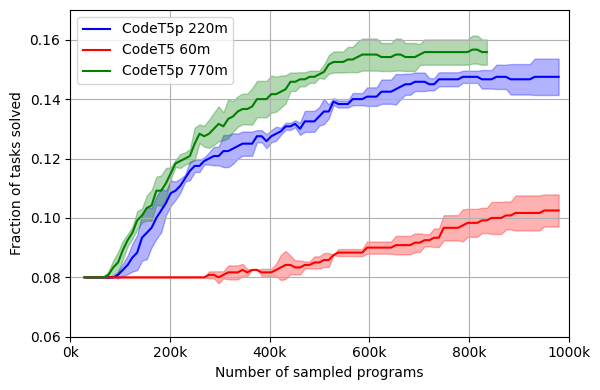}
    \caption{Cumulative performance as function of number of sampled programs for CodeIt with different model sizes}
    \label{fig:scaling}
\end{figure}

We perform a scaling study comparing CodeT5 model sizes and their effect on performance. Figure \ref{fig:scaling} shows a significant increase in performance increasing the model size from 60m to 220m and a smaller increase in performance from 220m to 770m. Note that for the 770m model, CodeIt was only run for 85 meta-iterations due to computation constraints.

It was our intention to add an even larger model to the scaling study (we opted for Mistral-7B). 
However, running CodeIt with this model took a lot longer (reaching 25 meta-iterations after 14 days) due to a the increased time needed for the sampling stage. 
We also found that replacing CodeIt's learning stage with LoRA-based finetuning (we carry out full-parameter finetuning with CodeT5) with a model of this size ultimately resulted in less than 9\% solved tasks after 200k sampled programs. For comparison, the CodeT5 220M model solves close to 11\%, and CodeT5 770M around 12\% at that same point. A possible reason is that such large models are more prone to overfitting given the small size of the dataset. We do believe LoRA-based finetuning of a larger model can be made to work, but that it would require a proper search of LoRA hyperparameters, as well as benchmarking different LMs in order to find the best backbone model for this task; we deem this to be beyond the scope of the present work.

\section{Program analysis}
\subsection{Failure Cases on Custom Validation Set}
\label{subsec:failure_cases}
\begin{figure}
    \begin{minipage}{0.51\linewidth}
        \centering
        \includegraphics[width=\linewidth]{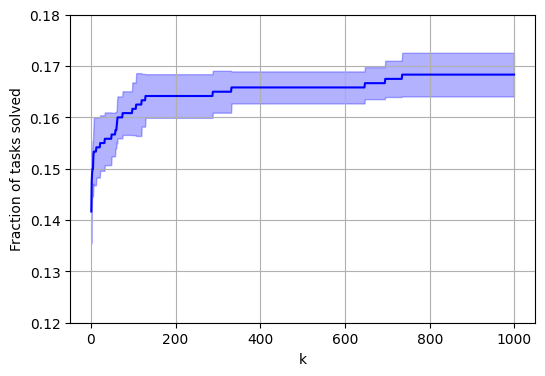}
        \caption{Pass@k curve for CodeIt over three seeds}
        \label{fig:pass@k}
    \end{minipage} \quad
    \begin{minipage}{0.47\linewidth}
        \centering
        \includegraphics[width=\linewidth]{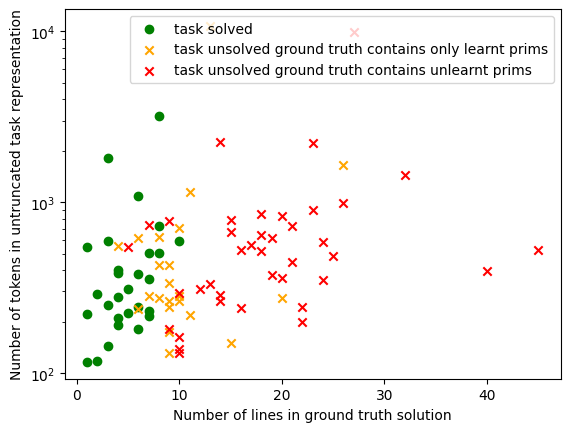}
        \caption{Failure cases on ARC custom validation set}
        \label{fig:fail_custom_val}
    \end{minipage}
\end{figure}

We analyze the properties of solved tasks on the custom validation set that we used for hyperparameter tuning. For this set, groundtruth programs are available, which allows us to analyze CodeIt's use of function primitives (the DSL operators). We ran CodeIt for 50 meta-iterations, and examined the program length and primitives of solved and unsolved tasks. Figure \ref{fig:fail_custom_val} visualises the results. We observe that CodeIt almost always solves the task if the solution program is short (10 lines or fewer). Further, CodeIt does not solve tasks of medium length (11 to 15 lines), but often learns to use their function primitives Finally, CodeIt never learns some primitives, especially those occurring in longer programs (16 lines or more). Perhaps unsurprisingly, CodeIt does not find a solution for tasks with a longer groundtruth program, as we can assume the tasks are more complex.
Future work could address this by including information on the execution state during program generation, effectively breaking complex problems up into subproblems.

\subsection{DSL Learning}
\label{subsec:primitives}
\begin{figure}
    \centering
    \includegraphics[width=\linewidth]{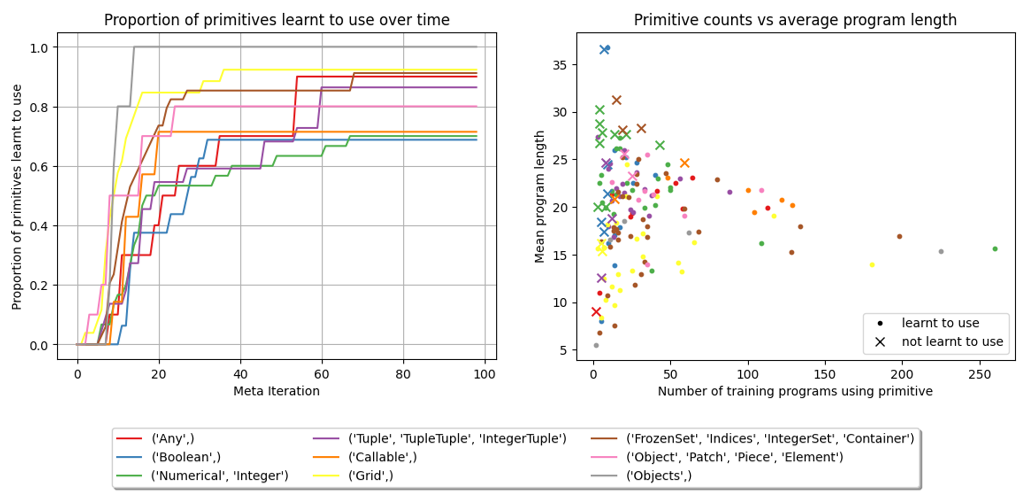}
    \caption{Primitives used in solutions for ARC evaluation set by output type. Left: proportion of primitives learnt over time. Right: scatter plot of counts vs mean program length. }
    \label{fig:fail_eval_set}
\end{figure}

We investigate how CodeIt learns to use DSL primitives on the ARC evaluation set. We plot the use of primitives over training time, and a scatter plot indicating for each primitive whether it is used, in Figure \ref{fig:fail_eval_set}. We find that some primitive types are learned quicker than others: for example, ones that return object representations are learned much quicker than ones that return numeric values. Moreover, primitives that are not learned typically occur in few training programs, or programs that are long, likely indicating it is more complex to learn to use these primitives.

\subsection{CodeIt compared with mutation baselines}
\label{appendix:codeitvsmutation}
\begin{table}[]
    \centering
    \begin{tabular}{l r}
        \toprule
        \textbf{Method} &  \textbf{Number of tasks solved}\\
        \hline
        CodeIt policy only & 23 \\
        Mutation $d_1$ only & 13 \\
        \hline
        CodeIt policy $\cap$ Mutation $d_1$ & 29 \\
        \bottomrule
    \end{tabular}
    \caption{ARC evaluation tasks solved per method. The top group of two rows show how many tasks were solved by a method, but not by the other. The final row shows tasks solved by both methods.}
    \label{tab: task_solved_per_method}
\end{table}
\begin{figure}
    \begin{minipage}{0.475\linewidth}
        \centering
        \includegraphics[width=\linewidth]{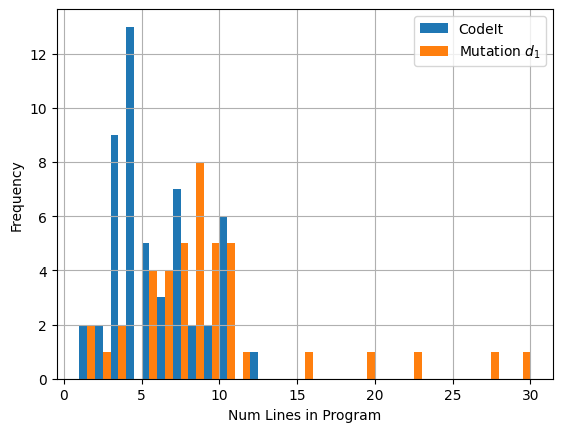}
        \caption{Histogram of number of lines for tasks where both CodeIt and Mutation produced solutions. CodeIt (in blue) produces shorter programs than the Mutation baseline (in orange).}
        \label{fig: program_length}
    \end{minipage} \quad
    \begin{minipage}{0.49\linewidth}
        \centering
        \includegraphics[width=\linewidth]{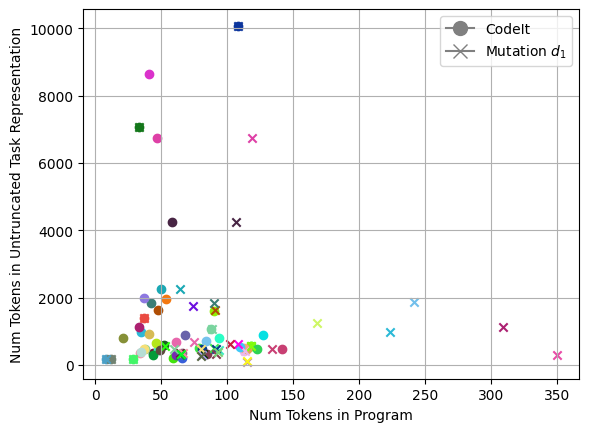}
        \caption{Number of task representation tokens vs number of program tokens. Colors represents the different tasks. We see no obvious correlation between task representation and program length.}
        \label{fig: program_length_task_rep}
    \end{minipage}
\label{fig: shortest_programs_per_method}
\end{figure}
We compare the programs found using our mutation $d_1$ baseline and the best performing of the three CodeIt runs. 
Table \ref{tab: task_solved_per_method} displays the number of ARC evaluation tasks uniquely solved by each method and the tasks which are solved by multiple methods. 
CodeIt's policy solves 52/400 tasks, 23 of which were not solved by the mutation baseline. 
In Figures \ref{fig: program_length} and  \ref{fig: program_length_task_rep}, we select the shortest program that solves an evaluation task for CodeIt and our mutation $d_1$ baseline, computing the program length and task representation size. Note that CodeIt has an encoder context window size of 1024 and so any tasks which having representations of more than 1024 tokens have been truncated. 
Overall, CodeIt finds shorter programs as shown in \ref{fig: program_length}. Further, for the same task, CodeIt more often finds shorter programs than our mutation $d_1$ baseline, as shown in \ref{fig: program_length_task_rep} where each color represents a different task. 
Interestingly, CodeIt does solve some tasks with very large task representations, suggesting in some cases a truncated task representation provides sufficient information to solve the task. 

In Table \ref{fig:comp_programs_short}, we show a subset of solution programs for ARC eval tasks solved by both CodeIt and our mutation $d_1$ baseline. 
We select tasks where the shortest programs differ between the two methods. CodeIt programs appear more concise and use different primitives. 
Out of the 29 tasks that are solved by both methods, there are 24 shortest programs where the programs differ by method. CodeIt only produces a longer program in 1 out of these 24 cases. The Mutation baseline often includes redundant lines.
In addition, for many programs, CodeIt produces a program that is qualitatively better: the solution is less complex, and contains fewer lines overall.

\begin{table}[]
    \centering
\begin{tabular}{llc}
\toprule
                                      CodeIt Policy &                                 Mutation $d1$  & Test Example \\  
\midrule                                  
                    \texttt{x1 = vmirror(I)} &                    \texttt{x1 = vmirror(I)} & \multirow{6}{*}{\includegraphics[width=4cm, keepaspectratio]{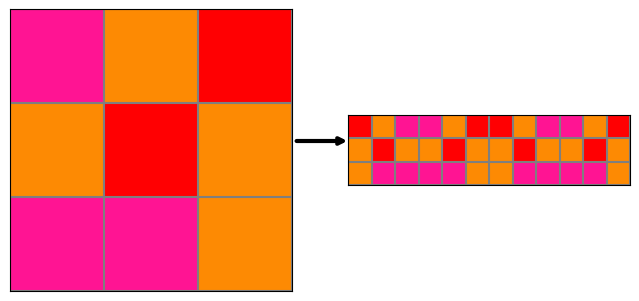}}\\
                \texttt{x2 = hconcat(x1, I)} &                \texttt{x2 = hconcat(x1, I)} \\
                \texttt{O = hconcat(x2, x2)} &                   \texttt{x3 = hmirror(x2)} \\
                                             &               \texttt{x4 = vconcat(x2, x3)} \\
                                             &               \texttt{x5 = hconcat(x3, x3)} \\
                                             &                    \texttt{O = hmirror(x5)} \\
\hline
                   \texttt{x1 = compress(I)} &                    \texttt{x1 = hmirror(I)} & \multirow{8}{*}{\includegraphics[width=4cm, keepaspectratio]{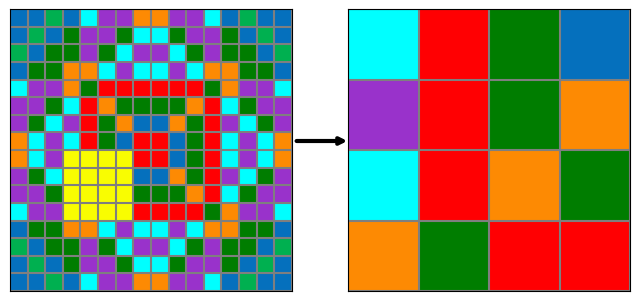}}\\
             \texttt{x2 = ofcolor(I, THREE)} &                    \texttt{x2 = vmirror(I)} \\
                     \texttt{x3 = rot90(x1)} &             \texttt{x3 = ofcolor(I, THREE)} \\
                \texttt{O = subgrid(x2, x3)} &               \texttt{x4 = subgrid(x3, x1)} \\
                                             &               \texttt{x5 = subgrid(x3, x2)} \\
                                             &                   \texttt{x6 = palette(x4)} \\
                                             &            \texttt{x7 = contained(ONE, x6)} \\
                                             &             \texttt{O = branch(x7, x5, x4)} \\
\hline
               \texttt{x1 = ofcolor(I, ONE)} &                  \texttt{x1 = mostcolor(I)} & \multirow{27}{*}{\includegraphics[width=4cm, keepaspectratio]{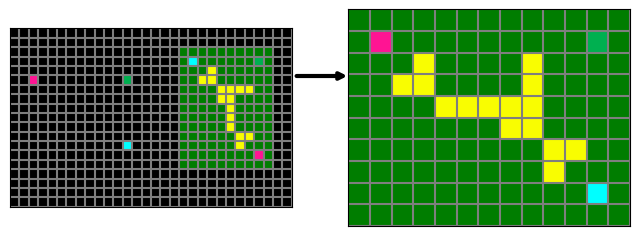}} \\
                \texttt{x2 = subgrid(x1, I)} &           \texttt{x2 = objects(I, T, F, T)} \\
                    \texttt{O = cmirror(x2)} &         \texttt{x3 = replace(I, x1, THREE)} \\
                                             &              \texttt{x4 = argmax(x2, size)} \\
                                             &              \texttt{x5 = argmin(x2, size)} \\
                                             &              \texttt{x6 = position(x4, x5)} \\
                                             &                     \texttt{x7 = first(x6)} \\
                                             &                      \texttt{x8 = last(x6)} \\
                                             &               \texttt{x9 = subgrid(x4, x3)} \\
                                             &                    \texttt{x10 = hline(x5)} \\
                                             &                  \texttt{x11 = hmirror(x9)} \\
                                             &                  \texttt{x12 = vmirror(x9)} \\
                                             &        \texttt{x13 = branch(x10, x11, x12)} \\
                                             &        \texttt{x14 = branch(x10, x7, ZERO)} \\
                                             &        \texttt{x15 = branch(x10, ZERO, x8)} \\
                                             &                \texttt{x16 = asobject(x13)} \\
                                             &        \texttt{x17 = matcher(first, THREE)} \\
                                             &           \texttt{x18 = compose(flip, x17)} \\
                                             &            \texttt{x19 = sfilter(x16, x18)} \\
                                             &                 \texttt{x20 = ulcorner(x4)} \\
                                             &                    \texttt{x21 = shape(x4)} \\
                                             &            \texttt{x22 = astuple(x14, x15)} \\
                                             &           \texttt{x23 = multiply(x21, x22)} \\
                                             &                \texttt{x24 = add(x20, x23)} \\
                                             &              \texttt{x25 = shift(x19, x24)} \\
                                             &                  \texttt{x26 = rot270(x11)} \\
                                             &                \texttt{O = paint(x26, x25)} \\
\hline
           \texttt{x1 = objects(I, F, F, T)} &           \texttt{x1 = objects(I, F, F, T)} & \multirow{5}{*}{\includegraphics[width=4cm, keepaspectratio]{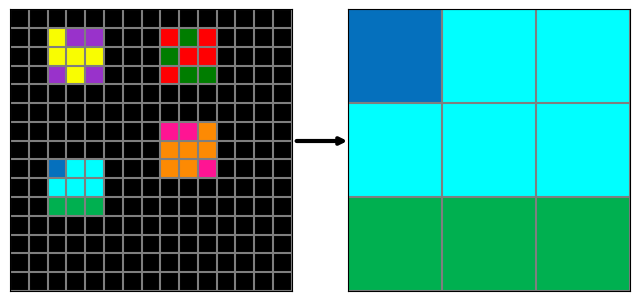}}\\
         \texttt{x2 = argmax(x1, numcolors)} &                 \texttt{x2 = leastcolor(I)} \\
                 \texttt{O = subgrid(x2, I)} &         \texttt{x3 = rbind(colorcount, x2)} \\
                                             &                \texttt{x4 = argmax(x1, x3)} \\
                                             &                 \texttt{O = subgrid(x4, I)} \\

\bottomrule
\end{tabular}
\caption{Selection of shortest programs for ARC evaluation tasks solved by CodeIt policy (left) and the Mutation $d_1$ baseline (right) for which CodeIt program is shorter.}
\label{fig:comp_programs_short}
\end{table}

\subsection{CodeIt over time}
\label{appendix:codeitvstime}
\begin{figure}
    \begin{minipage}{0.49\linewidth}
        \centering
        \includegraphics[width=\linewidth]{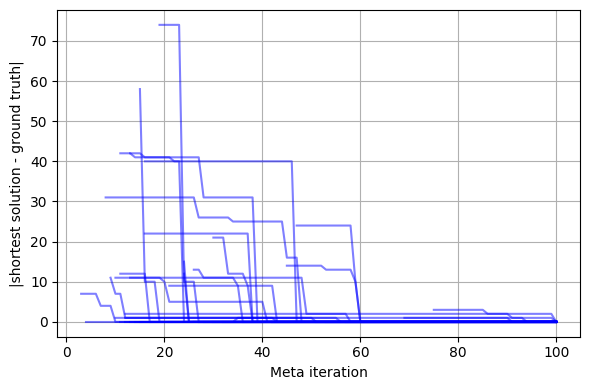}
        \caption{Difference in number of tokens between the shortest solution found per meta-iteration and shortest solution found by the final meta-iteration for best performing CodeIt run.}
        \label{fig: refinement}
    \end{minipage} \quad
    \begin{minipage}{0.49\linewidth}
        \centering
        \includegraphics[width=\linewidth]{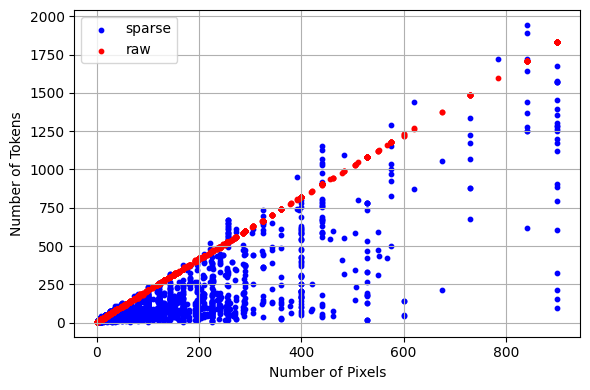}
        \caption{Grid size versus token count for the ARC training data. The sparse grid representation is typically shorter than the raw grid representation.}
        \label{fig: grid_tokens}
    \end{minipage}
\end{figure}

Since we do not have ground truth programs for the ARC evaluation set, we treat the shortest program found with demonstration performance and test performance equal to 1 for each task over all the meta-iterations as a proxy for the ground truth program. To examine how CodeIt solutions change over time, we take the subset of ARC evaluation tasks where the best performing CodeIt run finds such programs; this leaves us 45 tasks. We observe that once CodeIt finds a solution, CodeIt often continues to find both longer and shorter solutions in later meta-iterations. We pose that this gives the potential for program refinement, however, since the priority does not incorporate length, there is not explicit bias towards shorter solutions and so both longer and shorter solutions will be learned from. We observe that out of the 45 tasks, the best performing CodeIt run finds shorter solutions over time in 24 tasks as shown in Figure \ref{fig: refinement}.

In Tables \ref{tab:code_it_overtime}, we show a selection of examples where the best performing CodeIt run finds a longer solution in an earlier meta-iteration and shorter solution in a later meta-iteration.

\begin{table}[]
    \centering
\begin{tabular}{llc}
\toprule
                          Early Shortest Solution &                            Later Shortest Solution &                                                                                        Test Example \\
\midrule
         \texttt{x1 = ofcolor(I, EIGHT)} &    \texttt{x1 = replace(I, EIGHT, ZERO)} & \multirow{4}{*}{\includegraphics[width=4cm, keepaspectratio]{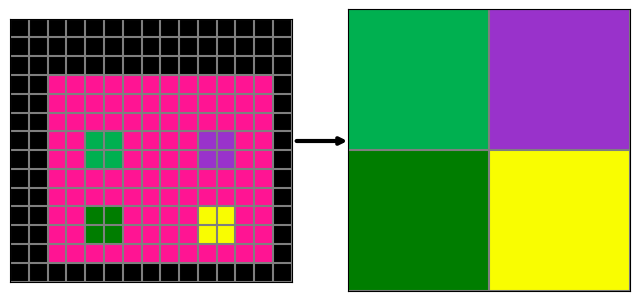}} \\
   \texttt{x2 = replace(I, EIGHT, ZERO)} &               \texttt{x2 = compress(x1)} &                                                                                                     \\
              \texttt{x3 = compress(x2)} &          \texttt{O = downscale(x2, TWO)} &                                                                                                     \\
         \texttt{O = downscale(x3, TWO)} &                                          &                                                                                                     \\
         & \\
\hline
       \texttt{x1 = objects(I, T, F, T)} &        \texttt{x1 = objects(I, T, F, T)} & \multirow{8}{*}{\includegraphics[width=4cm, keepaspectratio]{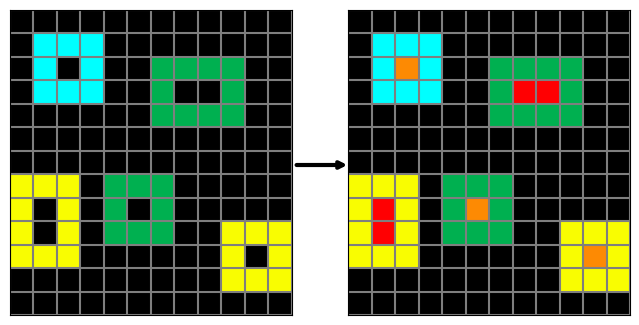}} \\
          \texttt{x2 = apply(delta, x1)} &           \texttt{x2 = apply(delta, x1)} &                                                                                                     \\
       \texttt{x3 = mfilter(x2, square)} &        \texttt{x3 = mfilter(x2, square)} &                                                                                                     \\
         \texttt{x4 = fill(I, FIVE, x3)} &          \texttt{x4 = fill(I, FIVE, x3)} &                                                                                                     \\
      \texttt{x5 = objects(x4, F, F, T)} &       \texttt{x5 = objects(x4, F, F, T)} &                                                                                                     \\
         \texttt{x6 = mapply(delta, x5)} &          \texttt{x6 = mapply(delta, x5)} &                                                                                                     \\
       \texttt{x7 = fill(x4, SEVEN, x6)} &         \texttt{O = fill(x4, SEVEN, x6)} &                                                                                                     \\
         \texttt{O = fill(x7, FIVE, x3)} &                                          &                                                                                                     \\
                                  \hline
       \texttt{x1 = objects(I, T, F, F)} &        \texttt{x1 = objects(I, T, F, T)} & \multirow{6}{*}{\includegraphics[width=4cm, keepaspectratio]{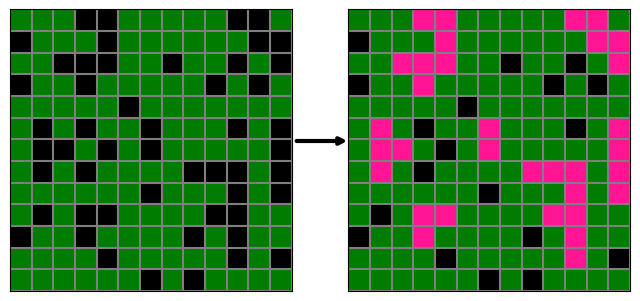}} \\
     \texttt{x2 = colorfilter(x1, ZERO)} &        \texttt{x2 = sizefilter(x1, ONE)} &                                                                                                     \\
       \texttt{x3 = sizefilter(x2, ONE)} &         \texttt{x3 = difference(x1, x2)} &                                                                                                     \\
        \texttt{x4 = difference(x2, x3)} &                  \texttt{x4 = merge(x3)} &                                                                                                     \\
                 \texttt{x5 = merge(x4)} &          \texttt{O = fill(I, EIGHT, x4)} &                                                                                                     \\
         \texttt{O = fill(I, EIGHT, x5)} &                                          &                                                                                                     \\
                                  \hline
                \texttt{x1 = vmirror(I)} &          \texttt{x1 = ofcolor(I, EIGHT)} & \multirow{5}{*}{\includegraphics[width=4cm, keepaspectratio]{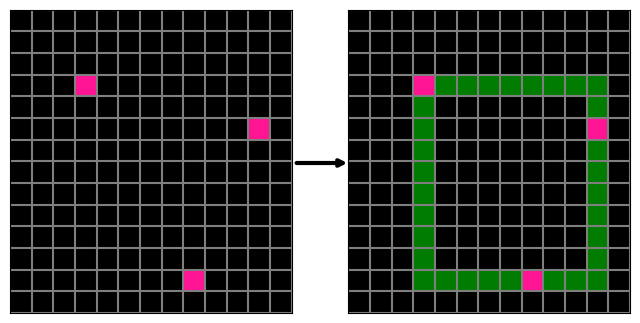}} \\
            \texttt{x2 = fgpartition(I)} &                    \texttt{x2 = box(x1)} &                                                                                                     \\
    \texttt{x3 = compose(outbox, inbox)} &       \texttt{O = underfill(I, ONE, x2)} &                                                                                                     \\
            \texttt{x4 = mapply(x3, x2)} &                                          &                                                                                                     \\
      \texttt{O = underfill(I, ONE, x4)} &                                          &                                                                                                     \\
\hline
\texttt{x1 = lefthalf(I)} &             \texttt{x1 = lefthalf(I)} &  \multirow{8}{*}{\includegraphics[width=4cm, keepaspectratio]{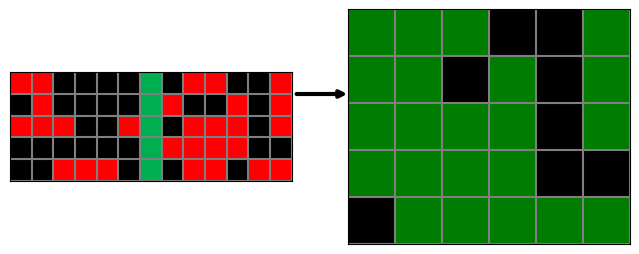}} \\
        \texttt{x2 = righthalf(I)} &            \texttt{x2 = righthalf(I)} &                                                                                                      \\
   \texttt{x3 = ofcolor(x1, ZERO)} &   \texttt{x3 = cellwise(x1, x2, ONE)} &                                                                                                      \\
   \texttt{x4 = ofcolor(x2, ZERO)} &  \texttt{O = replace(x3, SEVEN, ONE)} &                                                                                                      \\
\texttt{x5 = intersection(x3, x4)} &                                       &                                                                                                      \\
           \texttt{x6 = shape(x1)} &                                       &                                                                                                      \\
     \texttt{x7 = canvas(ONE, x6)} &                                       &                                                                                                      \\
   \texttt{O = fill(x7, ZERO, x5)} &                                       &                                                                                                      \\
                            \hline
         \texttt{x1 = lefthalf(I)} &             \texttt{x1 = lefthalf(I)} &  \multirow{6}{*}{\includegraphics[width=4cm, keepaspectratio]{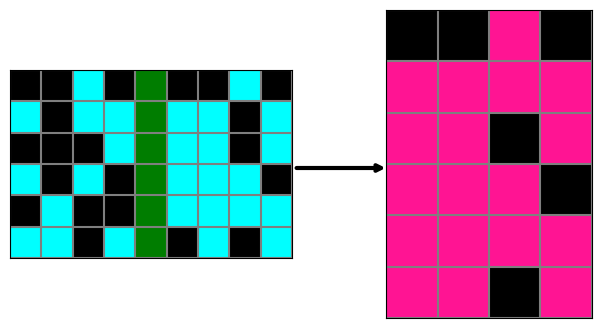}} \\
        \texttt{x2 = righthalf(I)} &            \texttt{x2 = righthalf(I)} &                                                                                                      \\
   \texttt{x3 = ofcolor(x1, FOUR)} &  \texttt{x3 = cellwise(x1, x2, FOUR)} &                                                                                                      \\
   \texttt{x4 = ofcolor(x2, FOUR)} & \texttt{O = replace(x3, FOUR, EIGHT)} &                                                                                                      \\
     \texttt{x5 = combine(x3, x4)} &                                       &                                                                                                      \\
  \texttt{O = fill(x1, EIGHT, x5)} &                                       &                                                                                                      \\
                            \hline
         \texttt{x1 = lefthalf(I)} &              \texttt{x1 = vmirror(I)} &  \multirow{8}{*}{\includegraphics[width=4cm, keepaspectratio]{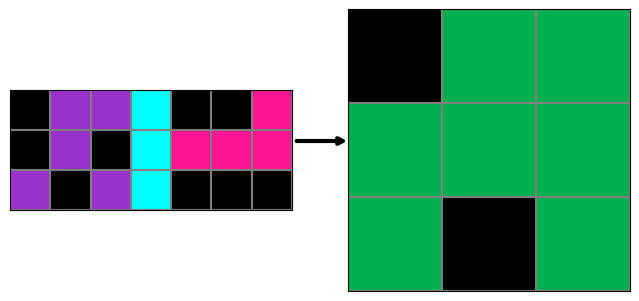}} \\
        \texttt{x2 = righthalf(I)} &             \texttt{x2 = lefthalf(I)} &                                                                                                      \\
   \texttt{x3 = ofcolor(x1, ZERO)} &            \texttt{x3 = righthalf(I)} &                                                                                                      \\
   \texttt{x4 = ofcolor(x2, ZERO)} &   \texttt{x4 = cellwise(x2, x3, TWO)} &                                                                                                      \\
\texttt{x5 = intersection(x3, x4)} &  \texttt{O = replace(x4, EIGHT, TWO)} &                                                                                                      \\
           \texttt{x6 = shape(x1)} &                                       &                                                                                                      \\
     \texttt{x7 = canvas(TWO, x6)} &                                       &                                                                                                      \\
   \texttt{O = fill(x7, ZERO, x5)} &                                       &                                                                                                      \\
\bottomrule
\end{tabular}
    \caption{Selection of shortest solutions for ARC evaluation tasks solved by CodeIt policy where shorter solutions are found over time.}
    \label{tab:code_it_overtime}
\end{table}

\section{ARC competitions}
\label{sec:arc_competitions}
\begin{table}[H]
    \centering
    \begin{tabular}{l l l c}
    \toprule
         Competition & Winner & Method & Hidden Test Perf. \\
         \midrule
         Kaggle 2020& \citet{kaggle}& Search in eval set DSL* & 21\%\\
         Kaggle 2020 late & Multiple \cite{kaggle_2020_leaderboard}& Ensemble previous entries* & 30\%\\
         ARCathon 2022& \citet{hodel} & Search in CodeIt DSL& 6\% \\
         ARCathon 2023& Multiple \cite{arc_2023_leaderboard}&  Unknown & 30\%  \\
    \end{tabular}
    \caption{Performance on Hidden Test Set for Various ARC Competition Winners. *Method conditions on ARC evaluation set.}
    \label{tab:arc_competitions}
\end{table}

\end{document}